\documentclass[a4paper, 10pt, twocolumn, twoside]{article}

\usepackage{ISARC}

\usepackage{lscape}
\usepackage{hologo}
\usepackage{xcolor}

\begin{document}

 % Do not change the following line
\linespread{0.5}

\title{Evaluation of Flight Parameters in UAV-based 3D Reconstruction for Rooftop Infrastructure Assessment}

\author{Nick Chodura$^{1, 2}$, Melissa Greeff$^{1, 3}$ and Joshua Woods$^{1, 2}$}

\affiliation{
$^1$Ingenuity Labs Research Institute, Queen's University, Canada\\
$^2$Department of Civil Engineering, Queen's University, Canada\\
$^3$Department of Electrical and Computer Engineering, Queen's University, Canada
}

\email{
\href{mailto:e.author1@aa.bb.edu}{16npc4@queensu.ca},
\href{mailto:e.author1@aa.bb.edu}{melissa.greeff@queensu.ca}, 
\href{mailto:e.author1@aa.bb.edu}{joshua.woods@queensu.ca}
}

% Do not change the following three lines
\maketitle 
\thispagestyle{fancy} 
\pagestyle{fancy}

\begin{abstract} 

Rooftop 3D reconstruction using UAV-based photogrammetry offers a promising solution for infrastructure assessment, but existing methods often require high percentages of image overlap and extended flight times to ensure model accuracy when using autonomous flight paths. This study systematically evaluates key flight parameters—ground sampling distance (GSD) and image overlap—to optimize the 3D reconstruction of complex rooftop infrastructure. Controlled UAV flights were conducted over a multi-segment rooftop at Queen’s University using a DJI Phantom 4 Pro V2, with varied GSD and overlap settings. The collected data were processed using Reality Capture software and evaluated against  ground truth models generated from UAV-based LiDAR and terrestrial laser scanning (TLS). Experimental results indicate that a GSD range of 0.75–1.26 cm combined with 85\% image overlap achieves a high degree of model accuracy, while minimizing images collected and flight time. These findings provide guidance for planning autonomous UAV flight paths for efficient rooftop assessments.

\end{abstract}


\section{Introduction}
\label{sec:Introduction}

The inspection of rooftops plays a critical role in assessing the structural integrity and maintenance needs of buildings. Traditionally, this task has been carried out by inspectors physically accessing rooftops to identify defects in the building envelope and rooftop equipment. This manual approach is inherently time-consuming, especially for large-scale commercial and industrial structures, and presents significant challenges for visualizing structures.

Advancements in Uncrewed Aerial Vehicle (UAV) technology and photogrammetry have opened new possibilities for building inspections. UAV's equipped with high-resolution cameras can capture images from locations inaccessible to inspectors, enabling comprehensive documentation of structures. These images can be processed using 3D reconstruction techniques to create digital models of the inspected structures. This not only facilitates better visualization of scale and defect locations, but also serves as a historical record for tracking structural changes over time.

\begin{figure}[t]
    \centering
    \includegraphics[trim=2.4cm 3.5cm 1.6cm 3.4cm, clip, width=0.48\textwidth]{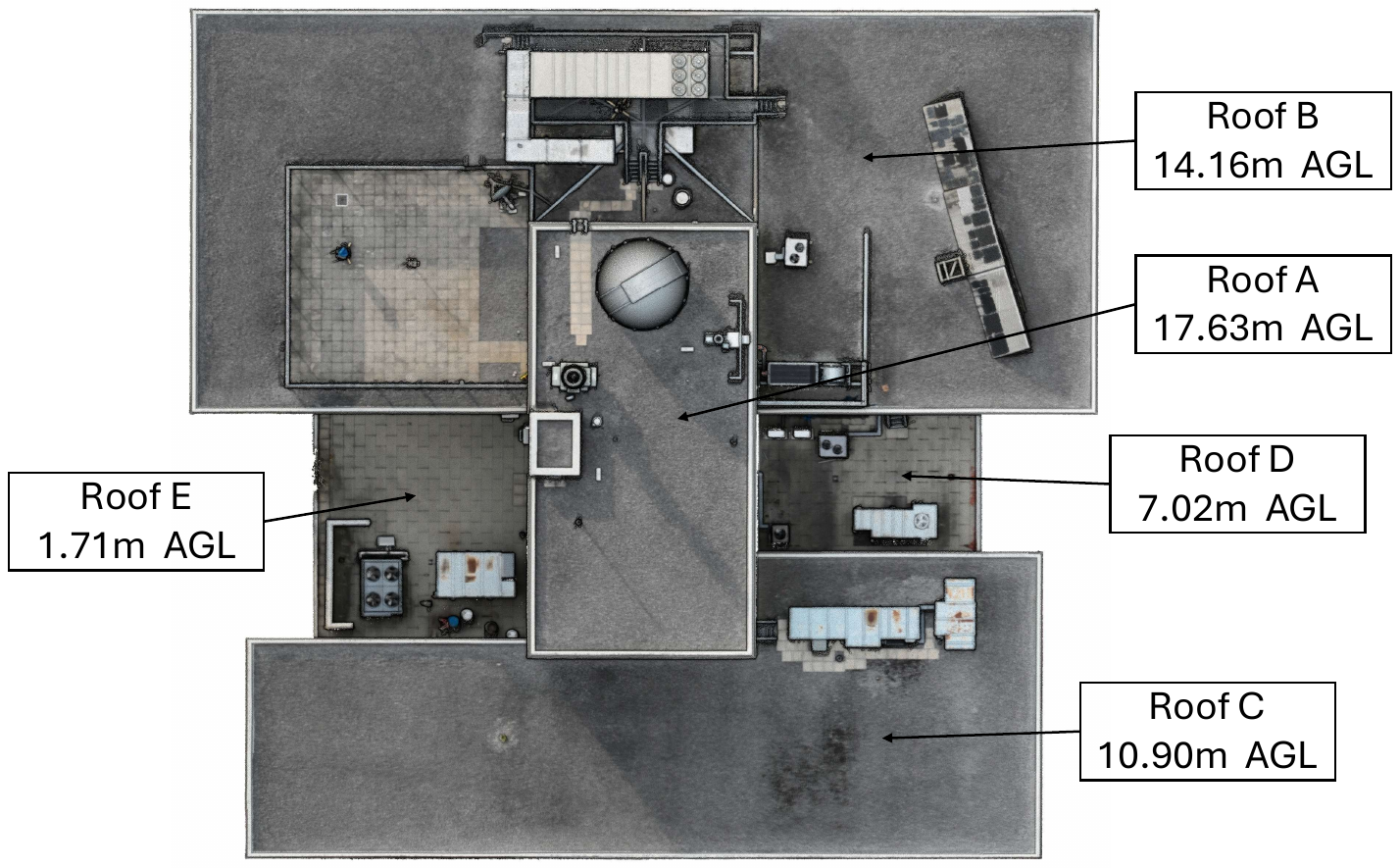}
    \caption{LiDAR point cloud of Ellis Hall, Kingston, Ontario with labeled roof sections and height above ground level (AGL).}
    \label{fig_gt}
\end{figure}

This study explores the use of autonomous flight paths for generating accurate 3D reconstructions for building inspections. Specifically, it investigates how varying flight parameters, such as image overlap (OL) and ground sampling distance (GSD), affect the accuracy of the reconstructed models. The intended outcome of this research is to be able to optimize these parameters to minimize the flight duration, computation time, and number of image captures - key factors for reducing inspection and processing time in commercial applications - while still maintaining accurate models. There are three primary contributions of this paper:

\begin{itemize}
\item
Recommendations based on experimental testing for optimal UAV flight parameters in autonomous photogrammetry-based rooftop 3D reconstruction.
\item 
Validation of a UAV-based LiDAR platform as a source of ground truth models using a high-resolution terrestrial laser scanner (TLS).
\item
Application of cloud-to-cloud comparisons and quantitative pointcloud analysis for infrastructure assessment applications.
\end{itemize}

\section{Related Work}
\label{sec:Related}

Related work was reviewed from two areas of research: (1) Influence factors for 3D reconstruction, and (2) Quantitative assessment of large-scale mesh models. 

\subsection{Influence factors for 3D reconstruction}

Influence factors for UAV photogrammetry-based 3D reconstruction can be broadly grouped into three main categories: flight parameters, camera settings/image acquisition, and reconstruction methods. While camera settings and reconstruction methods can have significant impacts on reconstruction quality, they were out of the scope of this research and were kept constant throughout the testing. The influence of flight parameters on photogrammetry reconstruction quality has been investigated thoroughly. Several studies have confirmed that flight height, image overlap, ground control point (GCP) quantities/distribution, and georeferencing methods have a significant impact on reconstruction accuracy. GCP’s are typically used in larger scale mapping applications such as forestry \cite{rs71013895, rs11101252} and glaciology \cite{rs9020186}, however there has also been research on GCP density and pattern on the reconstruction of dam infrastructure \cite{ZHAO2021103832}. In this paper's application of building facade 3D reconstruction, there are enough unique features from the building characteristics that an adequate number of tie points can be achieved during feature extraction without  the requirement of GCP’s. 

Flight height has been investigated individually \cite{anders2020, rs10060908}, and in combination with overlap and other parameters \cite{rs14164119, rs11080948, rs14164119, rs13040828, rs12142305, Xi2019} in various other applications, but less research has been done on how these parameters can influence the reconstruction quality of infrastructure assets. Altman et al. investigated how varying reconstruction methods, number of images, and number of control points (CP's) could influence the reconstruction quality of a church facade \cite{isprs-annals-IV-2-W4-199-2017}. However, all data collected was from terrestrial sources.  

\subsection{Quantitative assessment of mesh models}

Several methods have been established for quantitatively measuring model quality and accuracy. These methods can be broadly categorized into three areas: (1) Reconstruction metrics, (2) Discrete model metrics, and (3) Continuous model metrics. Metrics from the reconstruction process such as number of tie points and number of points in the dense cloud have been shown to correlate with lower control point RMSE values \cite{rs11101252}. Discrete metrics consist of control points with known locations that can be used to determine vertical and horizontal error at various locations on the model. These methods can be used to indicate drift or misalignment, but they do not provide continuous information about mesh accuracy across the entire model surface. Continuous model metrics can be divided into measures of mesh visual quality and measures of global geometric accuracy. Mesh visual quality metrics such as local roughness, noise, curvature variation, and topology can be used to quantify the appearance of a mesh surface \cite{s20205863}. Continuous metrics for measuring mesh accuracy are less common in large-scale 3D reconstruction due to the requirement of a ground truth model (GT), which can be difficult to capture at high degrees of coverage and accuracy. Typically, a GT is captured using TLS systems that can produce point clouds at a high degree of accuracy and point density \cite{isprs-annals-IV-2-W4-199-2017, Knapitsch2017, s20205863, bridge}. 

Chen et al. outlined a workflow for the inspection of a masonry bridge through UAV-based data acquisition, 3D reconstruction, data quality evaluation, and damage detection \cite{bridge}. They used a TLS point cloud to provide quality metrics for five different potential defects in the reconstructed photogrammetry model. They acknowledge the presence of occlusions in their TLS model due to obstacle-blocking and self-shadowing. These coverage-based defects would be exacerbated for roof reconstructions, where rooftop equipment and obstructions would limit the coverage of the TLS system to an even greater extent. In this paper, a two-stage GT method is proposed. A TLS is used on the rooftop to capture specific areas of high geometric complexity. This is used to validate a UAV-based LiDAR payload that collects data across the entire roof structure.

\section{Background}
\label{sec:Background}

Within the field of photogrammetry, the standard path for autonomous flights is a double-grid boustrophedon pattern using an oblique camera angle \cite{rs14164119}. This involves the UAV making back and forth passes at a specified distance in perpendicular directions across the coverage area. This method allows non-horizontal surfaces to be captured from four directions and reduces the chance of occlusions. These paths can be defined with two flight parameters: the image overlap (OL), and (b) the GSD. Because of the oblique camera angle and the varying roof elevations, GSD was used as a variable parameter instead of flight height to better represent the distance between the camera and the target surface.

GSD refers to the resolution of the captured images relative to the real-world dimensions of the scene. GSD depends on the distance between the camera and the object being captured  ($Z$), the camera focal length ($f$) and the size of a pixel on the sensor ($p$), which can be determined with the dimensions of the sensor and the image, i.e.,
\begin{equation}\label{eq_gsd}
\text{GSD} = \frac{Z}{f} \, p.
\end{equation}
OL considers the area of shared coverage between two consecutive images. This is expressed as a percentage of the entire image. Side OL can be determined using the distance between consecutive passes in the same direction. Front OL is defined as the following:
\begin{equation}\label{eq_overlap}
\text{OL} = 100 \cdot \frac{\text{GSD} \, w - v \, r_c}{\text{GSD} \, w},
\end{equation}
where $w$ is the image width, $v$ is the UAV speed, and \( r_c \) is the capture rate of the camera.

\begin{table*}[!h]
\centering
\caption{Variable Flight Parameters, Flight Data, and Model Data}
\begin{tabular}{lcccccc}
\hline
 & \multicolumn{2}{c}{Variable Parameters} & \multicolumn{2}{c}{Flight Data} & \multicolumn{2}{c}{Model Data} \\ 
\hline
Flight & GSD (cm) & OL (\%) & Duration (min) & Captures & Compute Time (min) & Tie Points \\ 
\hline
Flight 1 & 0.51 - 1.01 & 60 & 4 & 42 & 55 sec & 109 270 \\
Flight 2 & 0.51 - 1.01 & 70 & 6.5 & 76 & 1 min 26 sec & 637 177 \\
Flight 3 & 0.51 - 1.01 & 80 & 9.5 & 198 & 2 min 55 sec & 2 242 477 \\
Flight 4 & 0.51 - 1.01 & 85 & 15.75 & 249 & 5 min 7 sec & 1 148 312 \\
Flight 5 & 0.75 - 1.26 & 80 & 7.5 & 96 & 1 min 20 sec & 406 087 \\
Flight 6 & 0.75 - 1.26 & 85 & 13.5 & 233 & 3 min 31 sec & 4 571 844 \\
Flight 7 & 0.75 - 1.26 & 90 & 32.5 & 482 & 11 min 31 sec & 11 434 884 \\
Flight 8 & 0.98 - 1.49 & 85 & 8.25 & 171 & 3 min 11 sec & 3 073 278 \\
Flight 9 & 0.98 - 1.49 & 90 & 23.5 & 360 & 6 min 15 sec & 8 108 259 \\
\hline
\end{tabular}
\label{tab:flight_data}
\end{table*}

\section{Methodology}
\label{sec:Methodology}

In this study, the approach in assessing flight parameters for 3D photogrammetry has three steps: 1. Data Collection: collected image data from 9 flights of the building in Figure~\ref{fig_gt} using a DJI Phantom 4 Pro V2. The research team collected UAV-based LiDAR data using a DJI Matrice 350 RTK, and TLS data using a FARO Focus. 2. Model Generation: generated 3D reconstructions using photogrammetry software Reality Capture \cite{realitycapture2025}. LiDAR models are processed in DJI Terra \cite{dji_terra_2025}, and TLS models are aligned using SCENE point cloud software \cite{faro_scene_2025}. 3. Model Evaluation: All models were trimmed, aligned, and compared using 3D point cloud processing software CloudCompare \cite{cloudcompare2025}.

\subsection{Data collection}

Ellis Hall on the campus of Queen's University, Kingston, Ontario (latitude: 44.226, longitude: -76.496) was chosen as the test site for this experiment. The building features five distinct roof segments at AGL elevations ranging from 1.71 m to 14.16 m. The roof sections contain a variety of rooftop features, including an observatory dome, various HVAC equipment, fencing and barrier walls, and experimental testing equipment. This variance in elevation and subject matter provides unique levels of obstruction, lighting conditions, GSD, and feature complexity for the experiment. 

Images for the photogrammetry reconstructions were collected using a DJI Phantom 4 Pro V2, a consumer multirotor UAV with an on-board 20M camera featuring an 84° field of view and an 8.8 mm focal length, capable of capturing 4k images. The goal of this research was to assess the photogrammetry reconstruction quality of fine details and unique rooftop features in real-world conditions. As such, a process was needed that would provide a continuous, global metric of model accuracy across the entire mesh surface. To do this, A 'ground truth' model (GT) was needed to assess the accuracy of the generated reconstructions. A LiDAR scan of the building was performed using a DJI Matrice 350 RTK, equipped with a Zenmuse L2 LiDAR sensor. The L2 features a frame-based LiDAR, internal IMU system, and an RGB mapping camera \cite{dji_zenmuse_l2_specs}. A D-RTK 2 Base Station and on-board RTK Module provide Real-time Kinematic Processing (RTK) for accurate localization during the flight, and Post Processing Kinematics (PPK) was done using DJI Terra \cite{dji_terra_2025}. The L2 specifications report a ranging accuracy of 2 cm at 150 m scanning distance, which is reported to improve at closer ranges. This paper proposes the use of a point cloud collected from the Zenmuse L2 as a GT model to provide quantitative evaluations of accuracy for the photogrammetry-based 3D reconstructions. 

\begin{figure}[!htb]
    \centering
    \includegraphics[trim=8.5cm 9.3cm 7.6cm 8.2cm, clip, width=0.48\textwidth]{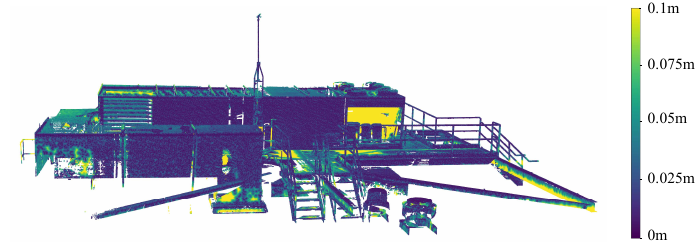}
    \caption{Recall scalar field for a cloud-to-cloud comparison of the UAV-based LiDAR and TLS point clouds.}
    \label{fig_vent}
\end{figure}

The UAV-based LiDAR point cloud was validated using the Faro Focus TLS platform. The TLS was placed in 14 locations across Roof Section B and Roof Section D. The Faro Focus produces a point cloud with 2 mm accuracy \cite{faro_focus_specs}. Ground control spheres were used to register the scans in Faro’s SCENE 3D point cloud software. Three test segments of the roof were chosen to compare the TLS and LiDAR point cloud's. Areas of geometric complexity were chosen to highlight potential deficiencies in the UAV-based LiDAR's effectiveness as a GT model. A cloud-to-cloud (C2C) comparison was conducted between the two point clouds at each test segment. One of these comparisons can be seen in Figure~\ref{fig_vent}. The test segments averaged an 85\% F-score at a threshold value of 2 cm.

\subsection{Model generation}

Photogrammetry models were generated using Reality Capture photogrammetry software \cite{realitycapture2025}. Reality Capture uses a photogrammetry pipeline consisting of feature extract, feature matching, and a Structure-from-Motion (SfM) algorithm to establish a set of tie points. A meshing stage generates a geometric surface for the scene using a triangulation algorithm. Finally, a Coloring/Texturing stage estimates a color for each vertex based on the input images.

Reality Capture's settings were kept to their default values with the exception of the following alignment parameters: the maximum features per megapixel were increased from 10,000 to 20,000 and the maximum features per image was increased from 40,000 to 100,000. The prior hardness of position and the prior hardness of orientation were decreased from 1.0 to 0.1. Preselector features were increased from 10,000 to 30,000 and the detector sensitivity was increased from medium to high.

\subsection{Model evaluation}

The models were evaluated using C2C distances between the photogrammetry-based models and the LiDAR point cloud. All point cloud analysis was done using CloudCompare \cite{cloudcompare2025}. The mesh models were sampled at 40,000 points per square meter. This was done to produce a dense point cloud that could accurately represent the mesh surface. All point clouds were then spatially subsampled with a minimum distance of 5 mm to ensure consistent uniform-density between models. The value of 5 mm was selected based on previous work on large-scale photogrammetry models \cite{Knapitsch2017}.

Models were then trimmed of surrounding surfaces and divided into five rooftop sections. The models were roughly aligned using equivalent point pairs, before a fine alignment was done using an iterative closest point (ICP) algorithm. C2C distances were calculated using a quadratic height function as a local surface model, with a nearest neighbour radius of 10 cm. Three metrics were produced, following procedures in \cite{Knapitsch2017}.

Precision is the measure of how accurately the points of the model align with the GT, represented as a percentage of points under a threshold distance $d$, and is defined as:
\begin{equation}\label{eq_precision}
P(d) = \frac{100}{|R|} \, \sum_{r \in R} [\mathrm{e}_{\mathrm{rg}} < d],
\end{equation}
where $R$ is the set of all points $r$ in the reconstruction. \( e_{rg} \) is the distance between a given point $r$ and its nearest point $g$ in the GT point cloud. 

Recall is a measure of the completeness of a reconstruction relative to a GT model. Recall is also represented as a percentage and is defined as:
\begin{equation}\label{eq_recall}
R(d) = \frac{100}{|G|} \, \sum_{g \in G} [\mathrm{e}_{\mathrm{gr}} < d],
\end{equation}
where $G$ is the set of all points $g$ in the GT point cloud.

The third metric combines these two measures to provide a robust qualifier for model quality that considers accuracy and completeness. The F-Score, or harmonic mean, is defined as:
\begin{equation}\label{fscore}
F(d) = \frac{2P(d)R(d)}{P(d) + R(d)}.
\end{equation}

\section{Results}

\begin{figure}[h!]
    \centering
    % Subfigure 1 (Top)
    \begin{subfigure}[!h]{\linewidth} % Use \linewidth to fit within a single column
        \centering
        \includegraphics[trim=2.9cm 7.3cm 16.4cm 6.2cm, clip, width=0.96\linewidth]{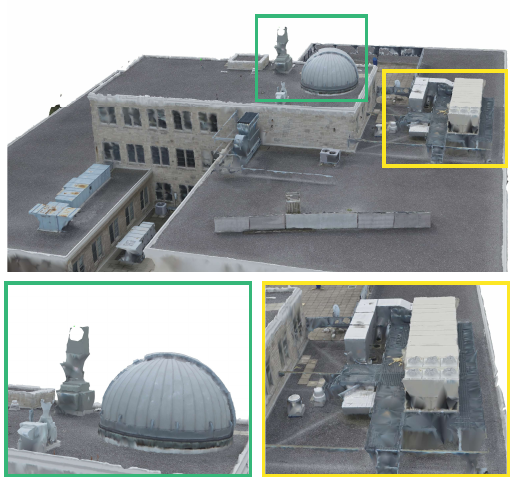}
        \caption{Flight 4: 0.51 - 1.01 cm GSD, 85\% OL}
        \label{fig:left}
    \end{subfigure}

    % Subfigure 2 (Bottom)
    \begin{subfigure}[!h]{\linewidth} % Use \linewidth to fit within a single column
        \centering
        \includegraphics[trim=11.8cm 7.3cm 7.5cm 6.1cm, clip, width=0.96\linewidth]{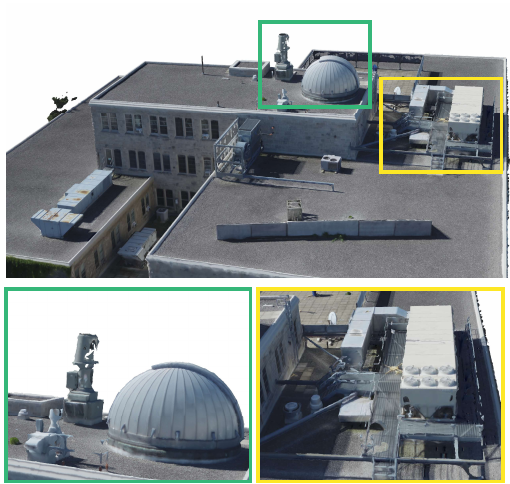}
        \caption{Flight 6: 0.75 - 1.26 cm GSD, 85\% OL}
        \label{fig:right}
    \end{subfigure}

    % Main caption
    \caption{At the same OL, Flight 6 produces a more accurate model than Flight 4. Thin-walled surfaces feature less occlusions, seen in the exhaust ventilator on the left, and complex geometry was represented better, seen in the walkway on the right.}
    \label{fig:comparison}
\end{figure}

\begin{figure*}[!h]
    \centering
    % Subfigure 1 (Top)
    \begin{subfigure}[h!]{\textwidth}
        \centering
        \includegraphics[trim=6.3cm 13.4cm 7.7cm 4.1cm, clip, width=0.96\textwidth]{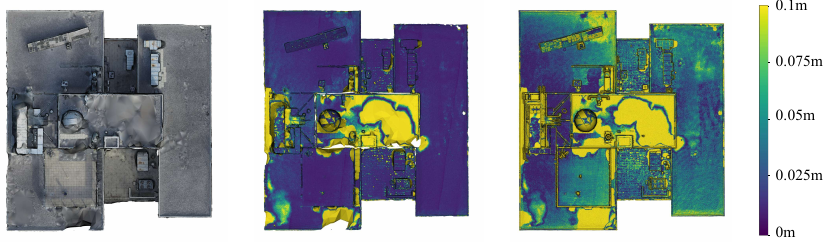}
        \caption{Flight 2: 0.51 - 1.01 cm GSD, 70\% OL}
        \label{fig:compare_top}
    \end{subfigure}

    % Subfigure 2 (Middle)
    \begin{subfigure}[h!]{\textwidth}
        \centering
        \includegraphics[trim=6.3cm 8.9cm 7.7cm 8.4cm, clip, width=0.96\textwidth]{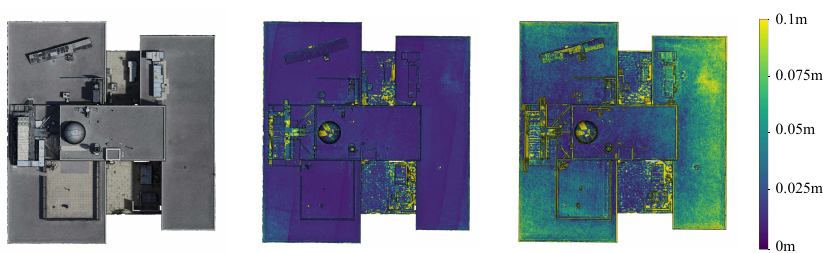}
        \caption{Flight 4: 0.51 - 1.01 cm GSD, 85\% OL}
        \label{fig:compare_mid}
    \end{subfigure}

    % Subfigure 3 (Bottom)
    \begin{subfigure}[h!]{\textwidth}
        \centering
        \includegraphics[trim=6.3cm 4.6cm 7.7cm 12.7cm, clip, width=0.96\textwidth]{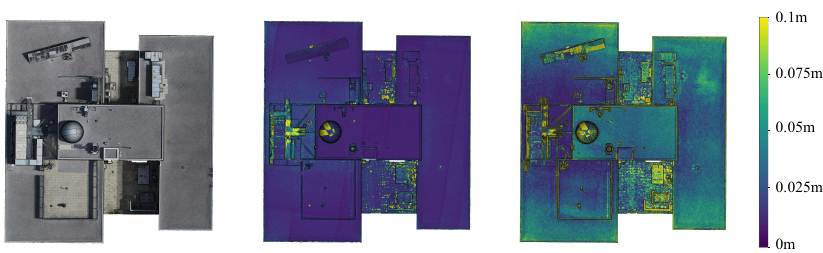}
        \caption{Flight 7: 0.75 - 1.26 cm GSD, 90\% OL}
        \label{fig:compare_bottom}
    \end{subfigure}

    % Main caption
    \caption{RGB (left), precision (middle), and recall (right) scalar fields for three of the test flights. The model from Flight 2 data produced large regions of high error (\textgreater 10 cm), and was unable to recreate simple building geometry. With higher overlap percentages, the error becomes primarily concentrated around regions of high detail, and the lower roof sections with poorer lighting and greater obstruction. Note that the rectangular area of high error on the observatory dome was due to rotation of the telescope prior to collection of the GT, however this is consistent across all test flights.}
    \label{fig:compare_full_page}
\end{figure*}

Based on qualitative observations, it was observed that the flights at 60\% and 70\% OL produced models with large distortions, as seen in Figure~\ref{fig:compare_top} and F-Scores in Table~\ref{tab:fscore}. At these overlap percentages, even some flat rooftop surfaces were not accurately represented by the model. Because of this observation, these OL percentages where not used in future testing. Above 80\% OL, models began to better represent the structure, but there were still variations in the reconstruction accuracy of fine details and complex geometry, as seen in Figure~\ref{fig:comparison}.

Some clear trends were observed from the C2C quantitative analysis conducted. The flights at 0.75 - 1.26 cm GSD generally outperformed the other flights, even when compared with flights that had significantly more images captured. This was observed for most roof sections: Flight 6 or Flight 7 had the highest F-score for four of the five roof sections. Another interesting observation is that Flight 5, having captured only 96 images, shared the third highest average rank with Flight 4 and Flight 9, which had 249 and 360 captures, respectively. It also had the lowest overlap of the three flights. The analysis also highlights the reduction in model quality for the lowest rooftop sections, D and E, at a higher GSD. At the same OL, Flight 9 had an F-Score that was 12.3\% lower for Roof E and 15.9\% lower for Roof D, when compared with Flight 7.

\begin{table*}[!h]
\centering
\caption{F-Scores at 4cm Threshold}

\begin{tabular}{lccccccccc}
\hline
 & Flight 1 & Flight 2 & Flight 3 & Flight 4 & Flight 5 & Flight 6 & Flight 7 & Flight 8 & Flight 9 \\
\hline
Roof A & 5.64 & 31.33 & 92.06 & 92.79 & 92.62 & \textbf{\textcolor{green!50!black}{95.82}} & \textbf{\textcolor{blue!80!green}{95.64}} & 94.89 & 95.51 \\
Roof B & 43.55 & 77.18 & 90.80 & 90.56 & 91.04 & \textbf{\textcolor{green!50!black}{92.39}} & \textbf{\textcolor{blue!80!black}{92.19}} & 90.98 & 91.56 \\
Roof C & 86.78 & 95.70 & 98.16 & 98.12 & 97.27 & \textbf{\textcolor{blue!80!green}{98.17}} & \textbf{\textcolor{green!50!black}{98.80}} & 97.43 & 98.01 \\
Roof D & 12.63 & 85.19 & 84.56 & 86.36 & \textbf{\textcolor{blue!80!green}{89.10}} & \textbf{\textcolor{green!50!black}{90.16}} & 88.53 & 70.46 & 84.36 \\
Roof E & 78.29 & 80.08 & 78.43 & \textbf{\textcolor{green!50!black}{88.92}} & \textbf{\textcolor{blue!80!green}{86.22}} & 86.03 & 82.24 & 76.67 & 84.01 \\
Mean & 29.16 & 69.13 & 88.54 & 91.26 & 91.18 & \textbf{\textcolor{green!50!black}{92.41}} & \textbf{\textcolor{blue!80!green}{91.30}} & 85.41 & 90.51 \\
Rank & 8.8 & 7.0 & 6.0 & 4.2 & 4.2 & \textbf{\textcolor{green!50!black}{1.6}} & \textbf{\textcolor{blue!80!green}{2.6}} & 6.4 & 4.2 \\
\hline
\end{tabular}
\label{tab:fscore}
\end{table*}

% put double here

\begin{figure*}[!htb]
    \centering
    \includegraphics[trim=0.8cm 3.55cm 1.5cm 11.73cm, clip, width=0.96\textwidth]{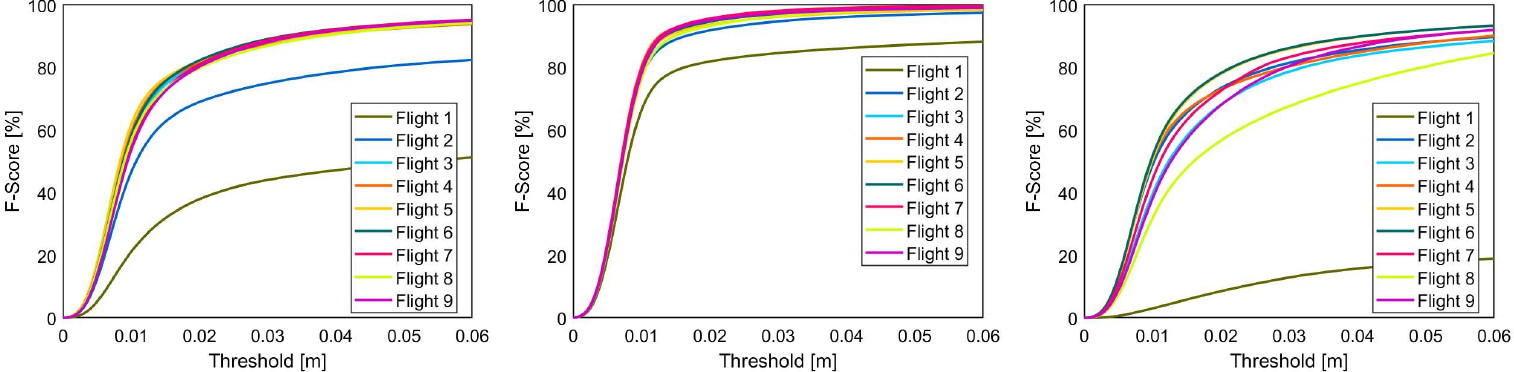}
    \caption{Precision at threshold values up to 6 cm for roof sections B, C, and D (left to right). For rooftops with less features, such as Roof C, the models remain fairly similar across different threshold values. As rooftop complexity is introduced, models begin to spread out.}
    \label{fig_1}
\end{figure*}

\begin{figure*}[!htb]
    \centering
    \includegraphics[trim=0.8cm 3.55cm 1.5cm 11.73cm, clip, width=0.96\textwidth]{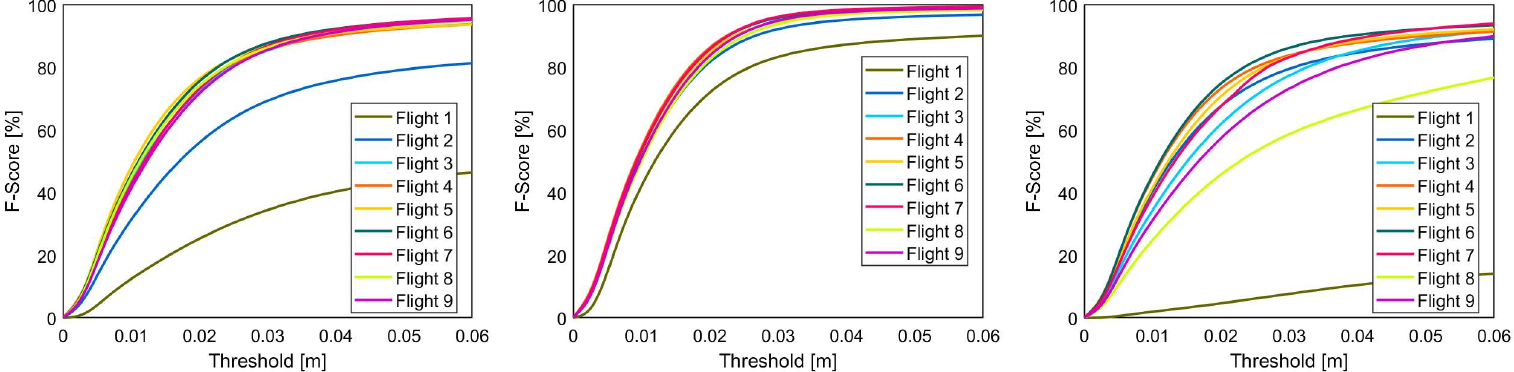}
    \caption{Recall at threshold values up to 6 cm for roof sections B, C, and D (left to right).}
    \label{fig_2}
\end{figure*}

The precision and recall were also plotted across different threshold values for each of the roof sections. In Figure 4 and Figure 5, an example of this is shown for roof sections B, C, and D. It is clear the flights at 0.75 - 1.26 cm GSD (gold, turquoise, and pink on the diagram) consistently perform well across a range of threshold values. These plots also illustrate the under-performance of the 0.98 - 1.49 cm GSD flights on Roof D, shown in lime green and purple on the graphs in Figure~\ref{fig_1} and Figure~\ref{fig_2}.

These figures also illustrate the separation in quality for more complex roof sections. Roof C, a section with minimal geometric complexity, features significantly less variation in F-Score across different flights. With the introduction of complex geometry (Roof B) and higher degrees of obstruction (Roof D), the flights begin to display a greater range in quality.

\section{Discussion}
\label{sec:Discussion}

The models produced with 90\% OL provided diminishing returns, or even decreasing quality, over the models at 85\% OL. It was also evident that the 0.75 – 1.26 cm GSD flights produced more accurate reconstructions relative to the number of images collected, particularly for the roof sections at higher elevations. For general roof surfaces that have a moderate amount of geometric complexity and limited obstructions, these flight parameters will provide high-quality mesh models that accurately represent the real-world structure.

This paper provides three additional findings with regard to autonomous flight paths for rooftop 3D reconstruction:

\begin{itemize}
    \item For roof sections that don’t exhibit complex geometry, lower OL can still yield effective results while reducing flight time, image captures, and computation time. This can be seen in Roof C, where 70\% OL produced a model with an F-score greater than 95\% at a 4cm threshold. 
    \item Thin-walled surfaces and fine details (e.g. chain-link fences, metal bar grating) were not captured accurately even at optimal OL and GSD values. Supplementary images would need to be captured manually around these features for accurate reconstruction.
    \item Lower roof sections exhibited a significant decrease in F-score relative to the other roof sections, regardless of GSD and OL parameters. This could be due to obstruction from other building sections and poorer lighting conditions. \\

\end{itemize}

\section{Future Work}
\label{sec:Future Work}
The number of flights was kept relatively low due to the time-intensive data collection and C2C analysis processes. In future testing, a larger sample size of flights should be compared. Other flight parameters have also been shown to have an effect on reconstruction quality, including ground control points, camera angle, image parameters, camera settings, and georeferencing methods \cite{rs14164119}. Future testing could investigate how these parameters may influence model quality in the context of building envelope 3D reconstruction. With added parameters and test flights, the main effects and interaction effects of these parameters can be observed using a statistic model.

The optimal parameters were also tested using other UAV platforms including the DJI Mavic 3 Enterprise and the DJI Mini 4 Pro. Flights were also conducted on other buildings, including residential and commercial structures. Future testing could involve quantitative evaluation for different UAV platforms and types of buildings.

\section{Conclusion}
\label{sec:Conclusion}

This paper investigated how varying flight parameters can influence the reconstruction accuracy of rooftop 3D models. Through a series of field tests, the study demonstrated that GSD values of 0.75 - 1.26 cm and an overlap of 85\% provided the most efficient flight path while maintaining a high degree of reconstruction accuracy. Using a rooftop TLS configuration, UAV-based LiDAR has been validated as an effective method for providing accurate rooftop point clouds that can be used as suitable ground-truth models for photogrammetry.

\section{Acknowledgments}
\label{sec:Acknowledgments}

This research was supported by Maintenance Drone Company and Mitacs through the Mitacs Accelerate Program. Additional support was provided by the Ingenuity Labs Research Institute at Queen's University.

\bibliography{ISARC}

\end{document}